\documentclass{article}
\usepackage{ijcai17}

\usepackage{times}
\usepackage{times}
\usepackage{epsfig}
\usepackage{graphicx}
\usepackage{amsmath}
\usepackage{amssymb}
\usepackage{graphicx}
\usepackage{epstopdf}
\usepackage{booktabs}
\usepackage{subfigure}

\usepackage{algorithm}
\usepackage{algorithmic}

\newtheorem{theorem}{Theorem}[section]

\newtheorem{proposition}[theorem]{Proposition}

\title{Revisiting $L_{2,1}$-norm Robustness with Vector Outlier Regularization}

\author{Bo Jiang\textsuperscript{a} \and Chris Ding\textsuperscript{b} \\
\textsuperscript{a}School of Computer Science and Technology, Anhui University, Hefei, China
\\ \textsuperscript{b}CSE Department, University of Texas at Arlington, Arlington, TX 76019, USA  \\ jiangbo@ahu.edu.cn, chqding@uta.edu}

\begin{document}

\maketitle

\begin{abstract}
In many applications, data usually contain outliers. One popular approach is to use $L_{2,1}$ norm function as a robust error/loss function.
 However, the robustness of $L_{2,1}$ norm function is not well understood so far.
 In this paper, we propose a new Vector Outlier Regularization (VOR) framework to understand
 and analyze the robustness of $L_{2,1}$ norm function.
Our VOR function defines a data point to be outlier if it is outside a threshold with respect to a theoretical prediction, and
regularize it --- pull it back to
the threshold line. We then prove that $L_{2,1}$ function is the limiting case of this VOR with the usual least square/$L_2$ error function as the threshold shrinks to zero.
One interesting property of VOR is that how far an outlier lies away from its theoretically predicted value
does not affect the final regularization and analysis results. This VOR property unmasks one of the most
peculiar property of $L_{2,1}$ norm function: The effects of outliers seem to be independent of how outlying they are ---
if an outlier is moved further away from the intrinsic manifold/subspace, the final analysis results do not change.
VOR provides a new way to understand and analyze the robustness of $L_{2,1}$ norm function.
Applying VOR to matrix factorization leads to a new VORPCA model. We give a comprehensive comparison with trace-norm based L21-norm PCA to demonstrate the advantages of VORPCA.
%
%
\end{abstract}

\section{Introduction}

Real-world image datasets  often contain noises and errors. Traditionally, this is often handled using Principal Component Analysis (PCA), Linear Discriminant Analysis (LDA), and
many other dimension reduction methods~\cite{LE,PatternClassification,NPE,LLE,MLDA}. Among the dimensionality reduction methods, PCA is one of most widely used linear algorithm because of their relative simplicity and effectiveness ~\cite{PatternClassification}. It assumes that the given high-dimensional data lie near in a lower-dimensional linear subspace. Given a dataset, the goal of PCA is to efficiently and accurately find this low-dimensional subspace. This problem can be efficiently solved by simply computing Singular Value Decomposition (SVD) on input data.
For Gaussian type noises, these methods are very effective. However, sometimes the noises are large, such as outliers, corrupted, occluded images, different illuminations, shading conditions, etc.
For these large noises or gross errors, PCA type  dimension reduction methods
usually break down.
 The robust dimension reduction or subspace extraction methods are developed for this purposes \cite{Torre:RPCA,Aanas:PAMI}.
 Some of the recent work use simple matrix norms such $L_1$ norm \cite{L1-PCA}, $L_{21}$-norm \cite{R1-PCA,RL1-PCA} to develop robust formulations.

 Although the above subspace learning methods are mainly {\it dimension reduction}, they simultaneously explicitly reduce the rank of the data. Recently, rank regularization (reduction) \cite{Cai:MatrixCompletion,Ma:TraceNorm} approaches have also been applied to reduce the rank of the data. All these studies use the trace norm as the main component for rank reduction \cite{Fazel:TraceNorm,Recht:TraceNorm}.
 In the $L_1$ norm based approach \cite{RPCA,Chandrasekaran:MIT}, authors shown the good effects for recovering true signals from large
 corruption.
 In another direction, the sparse subspace clustering/segmentation \cite{Elhamifar:Vidal,LLE} is studied with $L_{2,1}$ norm approach \cite{Liu:XXZL21Trace,Favaro:Vidal}.
 These rank regularization methods can correctly recover underlying low-rank structure in the data, even in the presence of noise.
 The main advantage of these trace-norm based rank reduction approach over earlier approaches is that the trace-norm is the convex envelop of rank of matrix and
 thus the optimization is convex. A unique optimal solution exist. The disadvantage is the computational speed: in many models,
 the augmented Lagrangian method is employed which envolves repeated SVD computation \cite{LinMaYi:AML,Favaro:Vidal}.

Over all, to deal with corrupted data, larger errors or outliers, the above methods generally use $L_1$ norm and $L_{2,1}$ norm to develop robust models.
However,  the robustness of $L_1$ and $L_{2,1}$ norm function are not well understood so far. In our previous work \cite{ORPCA}, we focus on scalar data and derive a model, called Outlier Regularization (OR), to deal with scalar outliers. Based on OR, we have presented a new analysis and explanation for $L_1$ norm robustness.
 But in many applications,  one need to deal with vector data, such as standard feature vectors in machine learning; for example, an images is usually represented by its feature vector, or sometimes, a vector of pixels.
In this paper, we focus on vector data and extend our previous outlier regularization to vector form.
We introduce a novel {\it vector outlier regularization} (VOR) function.
Although VOR function is a discrete function, it has an equivalent continuous representation.
We use VOR function to matrix factorization and propose Vector Outlier Regularization PCA (VORPCA).
VORPCA can be regarded as a balanced model between standard PCA and $R_1$-PCA \cite{R1-PCA,RL1-PCA} and degenerates to the $R_1$-PCA at the small tolerance limit. Using VORPCA, we provide a new intuitive analysis and interpretation for the  $L_{2,1}$ norm robustness.
%
%


\section{What is the $L_{2,1}$ norm robustness}

In this section, we explain the peculiar facts of $L_{2,1}$ error function robustness.
We show it from matrix factorization problem.
Formally, let $\textbf{X}=(\textbf{x}_1, \textbf{x}_2\cdots, \textbf{x}_n)\in \mathbb{R}^{p\times n}$  be the observed $n$ data points  in feature vector space.
Let $\textbf{U}\in \mathbb{R}^{p\times k}, k<p$ and $\textbf{V}=(\textbf{v}_1,\cdots \textbf{v}_n)\in \mathbb{R}^{k\times n}$,
%
we consider two types of low-rank matrix factorization problems \cite{R1-PCA,PatternClassification},
\begin{equation}\label{EQ:L2}
\min_{\textbf{U},\textbf{V}} \  \ E_2(\textbf{U},\textbf{V}) = \sum^n_{i=1} \|\textbf{x}_i - \textbf{U}\textbf{v}_i\|^2 = \|\textbf{X} - \textbf{U}\textbf{V}\|^2_F
\end{equation}
\begin{equation}\label{EQ:L2}
\min_{\textbf{U},\textbf{V}} \  \ E_{21}(\textbf{U},\textbf{V}) = \sum^n_{i=1} \|\textbf{x}_i - \textbf{U}\textbf{v}_i\| = \|\textbf{X} - \textbf{U}\textbf{V}\|_{2,1}
\end{equation}
where $\|\cdot\|$ is the $L_2$ norm of vector.
The traditional intuitive understanding of $L_{2,1}$ robustness is follows:

Suppose $\textbf{x}_{i_1}$ is an outlier, then the residual $r_{i_1} = \|\textbf{x}_{i_1} - \textbf{U}\textbf{v}_{i_1}\|$ is larger than residuals of other vector data points.
In $E_2(\textbf{U},\textbf{V})$, due to the squaring, $r^2_{i_1}$ would be much larger than other squared residuals and thus easily dominate the objective function.
In $E_{21}(\textbf{U},\textbf{V})$, the error for each data point is $r_{i_1} = \|\textbf{x}_{i_1} - \textbf{U}\textbf{v}_{i_1}\|$, which is not squared, and thus  diminishes the undue influence of those outliers and thus makes the learning more robust or stable.

However, we think this understanding of $L_{2,1}$ robustness is questionable.
Although the large errors due to outliers are not squared in $L_{2,1}$, they are still large and thus one would expect they would still significantly influence the cost function and therefore the final results.
In fact, experimental results demonstrate that in $L_{2,1}$ the outliers have \emph{small influence} on the final results\footnote{However, this is different from throwing the outliers out. The VOR process shows that the effect of an outlier is almost the same as a data point on the threshold line,
which is not zero. }.
In other words, $L_{2,1}$ function is insensitive w.r.t. outlyingness of the outliers: as long as a data point $\textbf{x}_{i_1}$ is an outlier, how far away $\textbf{x}_{i_1}$ lies does not affect the final results. This insensitivity property is one of the most peculiar property of $L_{2,1}$ norm; $L_1$ norm has
the same insensitivity property~\cite{ORPCA}.

%
\begin{figure}[!htb]\label{L21Robustness}
\centering
\includegraphics[scale=0.43]{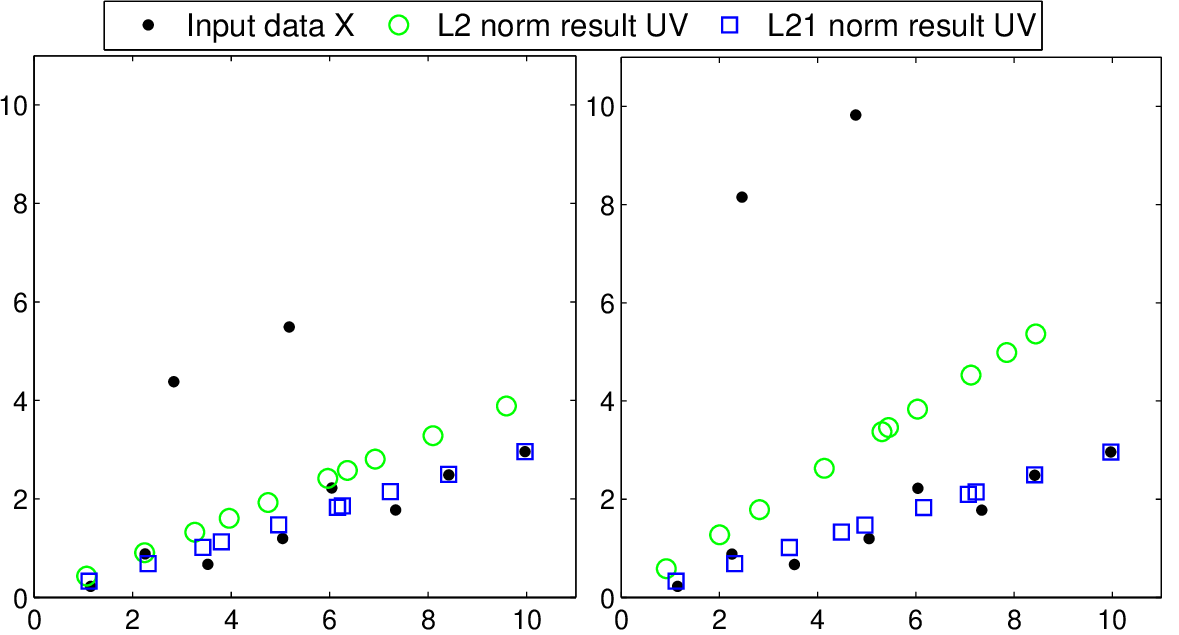}
\caption{Illustration of $L_{2,1}$ robustness on toy dataset.
In each panel, 10 data points are generated, two of which are outliers.
For each data point, we use $\textbf{U}\textbf{v}_i$ to fit/predict each data point $\textbf{x}_i$. }
\end{figure}
\begin{figure}[!htb]\label{L21Robustness}
\centering
\includegraphics[scale=0.175]{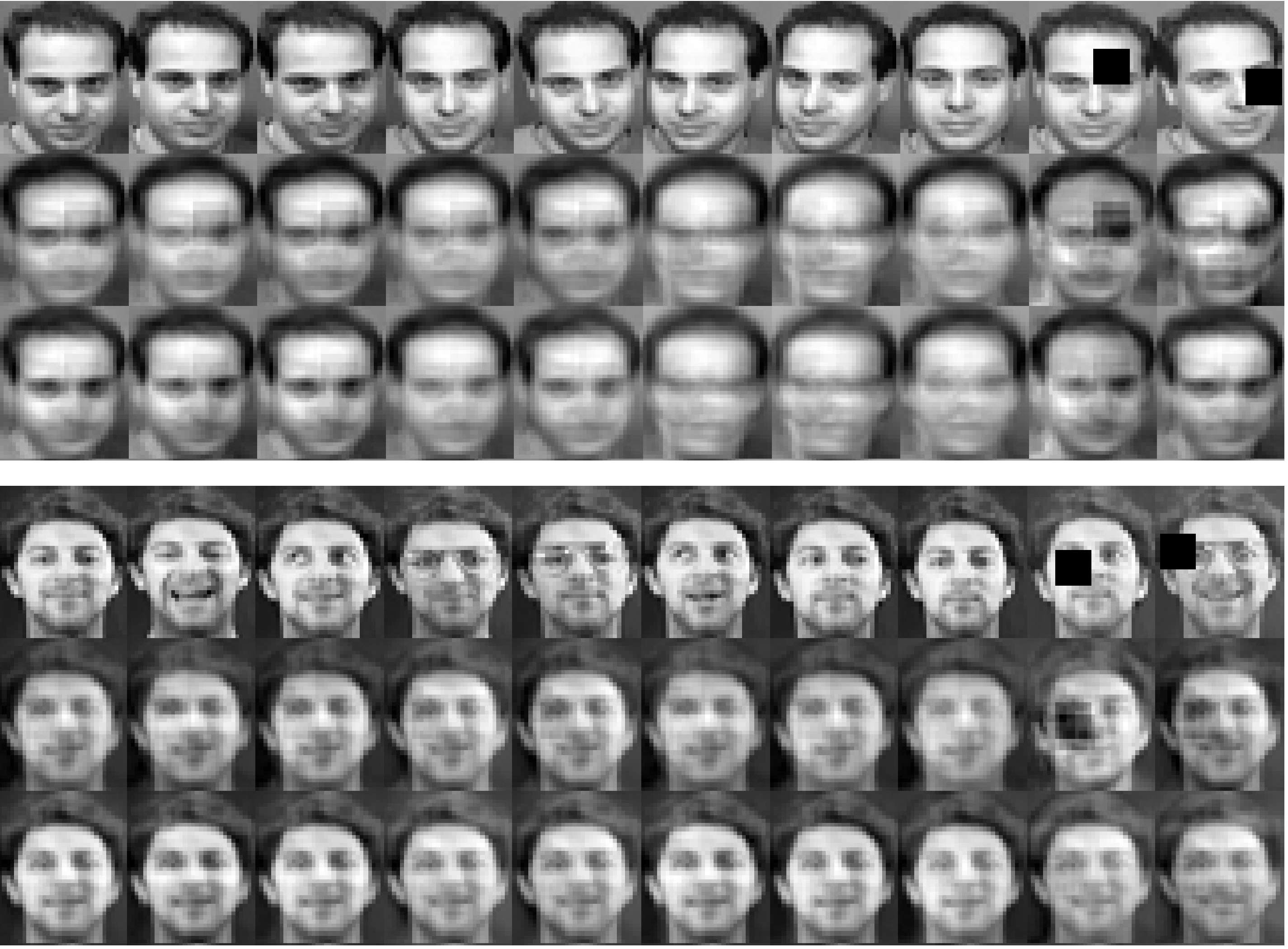}
\caption{Illustration of $L_{2,1}$ robustness on AT\&T face data.
In each panel, top row: original input images; middle row: $L_2$ reconstruction $\textbf{U}\textbf{V}$; bottom: $L_{2,1}$ reconstruction $\textbf{U}\textbf{V}$. }
\end{figure}

Figure 1 and 2 show some examples on both toy data and AT\&T face data, respectively.
In Fig. 1, we use 2D toy data to show the robustness of $L_{2,1}$ norm (Eq.(2)).
Note that, the $L_2$ results are
heavily influenced by outliers; but the $L_{2,1}$ results do not seem to be influenced by the outliers. In fact, the
outliers in the two datasets (left and right panels) are different; But $L_{2,1}$  results on these two datasets are exactly same, i.e., the outliers do not seem to influence the $L_{2,1}$ results. Note that the errors due to outliers (in the $L_{2,1}$ function) are much larger than those from non-outliers. 
Figure 2 shows the reconstruction results on AT \&T face data (100 images of 10 persons). Two images of each person are selected and corrupted to generate outlier images.
More details are given in Experimental section.
Note that, the $L_{2,1}$ performs more robustly than $L_2$ results when outliers exist. 

In the remaining of this paper, we show that this $L_{2,1}$ robustness property is due to a process of \textbf{vector outlier regularization}.
The  $L_{2,1}$  function minimization is the limiting case of vector outlier
regularization. 

\section{Vector Outlier Regularization Function}



In this section, we  propose our vector outlier regularization (VOR) function.

Let $\textbf{X}=(\textbf{x}_1, \textbf{x}_2\cdots, \textbf{x}_n)\in \mathbb{R}^{p\times n}$  be the observed input data  in feature vector space.
Let $\textbf{F}=(\textbf{f}_1, \textbf{f}_2\cdots, \textbf{f}_n)\in \mathbb{R}^{p\times n}$ be the corresponding theoretical model prediction.
%
We define a vector data point $\textbf{x}_i$ to be significantly distorted or highly corrupted
if the difference $\| \textbf{x}_i - \textbf{f}_i \|$  between the observed measured data $\textbf{x}_{i}$ and the theoretical prediction $\textbf{f}_{i}$
is bigger than a tolerance limit $\delta$.
%
We wish to {\bf correct} these highly corrupted data points.
One intuitive and effective way is to move them
towards the prediction manifold, but keep them at the boundary (tolerance limit). We achieve this purpose by defining the following function, 

%
\begin{equation}\label{EQ:robust-function}
\tilde{\textbf{x}}_{i} =  \left\{
  \begin{array}{l l}
    \textbf{x}_{i}                                                    & \; \text{if}\;  \|\textbf{x}_{i} - \textbf{f}_{i}\| \le \delta  \\
    \textbf{f}_{i} + \delta \dfrac{\textbf{x}_{i} - \textbf{ f}_{i}}{\|\textbf{x}_{i} - \textbf{ f}_{i}\|} & \; \text{if}\;  \|\textbf{x}_{i} - \textbf{f}_{i}\| > \delta\\
  \end{array} \right.
\end{equation}
where $\delta > 0$ is a threshold/tolerance parameter.
$\|\textbf{v} \|$ is the Euclidean norm of vector $\textbf{v}$.
We call it {\bf vector outlier regularization} (VOR) function.
Figure 3 shows an illustration of VOR function. 

%
\begin{figure}[!htb]\label{corruptionTolerant}
\centering
\includegraphics[scale=0.375]{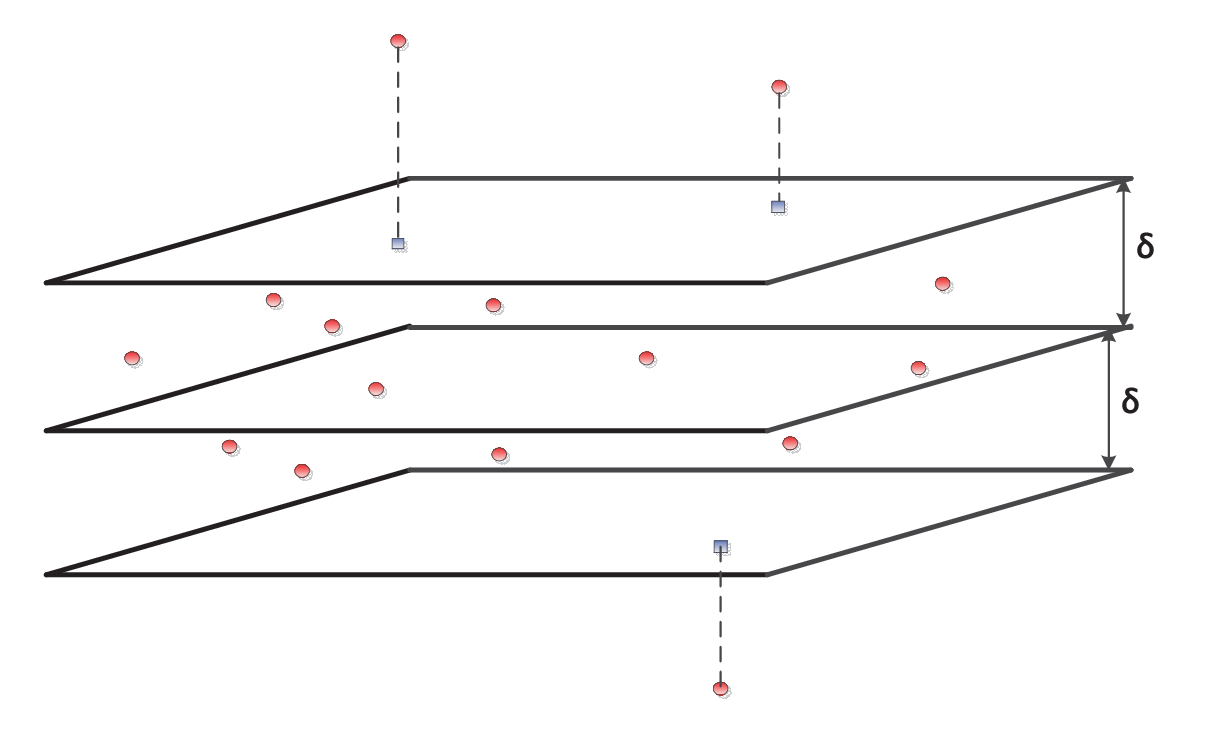}
\caption{Illustration of vector outlier regularization function of Eq.(3).
Red points $\textbf{x}_1 \cdots \textbf{x}_n$  are input data. Their theoretical predictions $\textbf{f}_1 \cdots \textbf{f}_n$ are on the plane(the subspace).
Top and bottom planes
indicate threshold planes of tolerant limits: data points $\textbf{x}_i$ outside threshold plane are considered outliers; they are pulled back to the threshold plane $\tilde{\textbf{x}}_i$ by ``outlier regularization".
Data points within tolerant limits remain unchanged.}
\end{figure}
%

%
One main feature of the above VOR is that an outlier $\textbf{x}_i$ is defined against a theoretical prediction $\textbf{f}_i$, whereas in
standard statistics, an outlier is defined against the rest of the input data. Our VOR is not the usual outlier removal and it aims to provide a sensible definition of outlier and then correct (reconstruct) them according to theoretical prediction. 


\subsection{Continuous representation}
The VOR function Eq.(\ref{EQ:robust-function}) is discrete. It is difficult to efficiently compute this discrete function when actually solving a large problem. Fortunately, our VOR  also has an equivalent continuous variational representation. We have the following,
\begin{proposition}
  $\widetilde{\textbf{X}}=(\tilde{\textbf{x}}_1, \tilde{\textbf{x}}_2,\cdots \tilde{\textbf{x}}_n) \in \mathbb{R}^{p\times n} $ of Eq.(\ref{EQ:robust-function})
    is the optimal solution to
the following optimization problem, i.e.,
\begin{equation}\label{EQ:robust-function-Con}
\widetilde{\textbf{X}}  = \arg\min_{\textbf{Z}} \  \|\textbf{X}-\textbf{Z}\|_{2,1} + {1\over 2\delta} \|\textbf{Z}-\textbf{F}\|_F^2.
\end{equation}
\end{proposition}
%
\textbf{Proof.}
Due to the $L_{2,1}$-norm, the formulation Eq.(\ref{EQ:robust-function-Con})
can be decoupled into $n$ separate independent sub-problems:
\begin{equation}\label{EQ:robust-function3}
\tilde{\textbf{x}}_i  = \arg\min_{\textbf{z}_i} \  \|\textbf{x}_i-\textbf{z}_i\|+ {1\over 2\delta} \|\textbf{z}_i -\textbf{f}_i\|^2.
\end{equation}
We now prove that the solution of Eq.(\ref{EQ:robust-function3}) is given by Eq.(1).
Setting $ \textbf{u} = \textbf{z}_i - \textbf{x}_i $, Eq.(\ref{EQ:robust-function3}) can be written as
\begin{equation}\label{EQ:proximal}
\min_{\textbf{u}} \   \delta\| \textbf{u} \| + \frac{1}{2} \|\textbf{u} -(\textbf{f}_i - \textbf{x}_i) \|^2.
\end{equation}
In Appendix, we prove that the solution of
Eq.(\ref{EQ:proximal})
  is given by
\begin{equation}
\textbf{u}^* = \max \big( 1 - \frac{\delta}{\|\textbf{f}_i - \textbf{x}_i\|}, 0 \big) \big( \textbf{f}_i - \textbf{x}_i \big)  .
\label{EQ:robust-function4}
\end{equation}
Thus for Eq.(\ref{EQ:robust-function3}),  $\tilde{\textbf{x}}_i = \textbf{z}_i^* = \textbf{u}^* + \textbf{x}_i$.\\
If  $\|\textbf{f}_i - \textbf{x}_i\| \le \delta$,
$\textbf{u}^* = \textbf{0}$, thus $\textbf{z}_i^* = \textbf{x}_i$ which is the same as
Eq.(\ref{EQ:robust-function}).\\
If  $\|\textbf{f}_i - \textbf{x}_i\| >  \delta$, we have
$$
\textbf{u}^* = \big(1 - \frac{\delta}{\|\textbf{f}_i -\textbf{x}_i\|}\big) \big( \textbf{f}_i - \textbf{x}_i \big)
= \textbf{f}_i - \textbf{x}_i  -  \delta \frac{\textbf{f}_i - \textbf{x}_i}{\|\textbf{f}_i - \textbf{x}_i\|}.
$$
Thus
$ \textbf{z}_i^*
= \textbf{u}^* + \textbf{x}_i = \textbf{f}_i   + \delta \dfrac{\textbf{x}_i - \textbf{f}_i}{\|\textbf{x}_i - \textbf{f}_i\|},
$
which is the same as
Eq.(\ref{EQ:robust-function}). This completes the proof.
$\hfill \Box$
%

\section{Vector Outlier Regularization in Matrix Factorization}

Here we apply the VOR function to matrix factorization.
Let $\textbf{X}=(\textbf{x}_1, \cdots, \textbf{x}_n)\in \mathbb{R}^{p\times n}$  be the observed input image data.
Let $\textbf{F}=(\textbf{f}_1, \cdots, \textbf{f}_n)\in \mathbb{R}^{p\times n}$ be the corresponding theoretical prediction.
Here, we set the theoretical prediction model to be the rank-$k$ approximation same as PCA, i.e., $\textbf{F} = \textbf{U}\textbf{V}$, where $\textbf{U}\in \mathbb{R}^{p\times k}, \textbf{V}\in \mathbb{R}^{k\times n}$ and $\mathrm{rank}(\textbf{U}\textbf{V})\leq k$.
Then, we aim to solve, 
\begin{align}\label{EQ:ORPCA0}
& \min_{\tilde{\textbf{X}} , \textbf{U},\textbf{V}} \  \|\tilde{\textbf{X}}  -\textbf{U}\textbf{V}\|^2_F \\
&  s.t. \; \{\tilde{\textbf{x}}_i, \textbf{f}_i = (\textbf{U}\textbf{V})_{i}\} \; \text{satisfy Eq.(\ref{EQ:robust-function}})
\end{align}
where $\tilde{\textbf{X}} = (\tilde{\textbf{x}}_1, \tilde{\textbf{x}}_2,\cdots \tilde{\textbf{x}}_n) \in \mathbb{R}^{p\times n}$  and $ (\textbf{U}\textbf{V})_i$ is the $i$-th column of $\textbf{U}\textbf{V}$.
In this paper, we call it as Vector Outlier Regularization PCA (VORPCA).
Note that $\tilde{\textbf{x}}_i$ provides a kind of reasonable reconstruction for the input data $\textbf{x}_i$ and also has some important properties. We will discuss it in the following section in detail. In the following, we first provide an intuitive illustration of VORPCA and derive an effective algorithm to solve VORPCA problem. 
\\

\noindent  \textbf{Illustration}. In Figure 4, we show the results of VORPCA on a simple 2D data set. The original data $\{\textbf{x}_i\}$  are shown in black dots. Reconstructed data $\{ \tilde{\textbf{x}}_i \}$  are shown as red-circles(non-outliers) and blue-squares (outliers). Red line indicates prediction correct subspace (standard PCA on the reconstructed data $\{\tilde{\textbf{x}}_i\}$), while green lines show the boundary (tolerance limit). Outliers are brought back to the correct
subspace by VORPCA at several $\delta$ values but kept at the boundary (tolerant limit) at convergence. 
\begin{figure}[!htb]\label{2DAPP-L21Reg-PCA}
\centering
\includegraphics[scale=0.475]{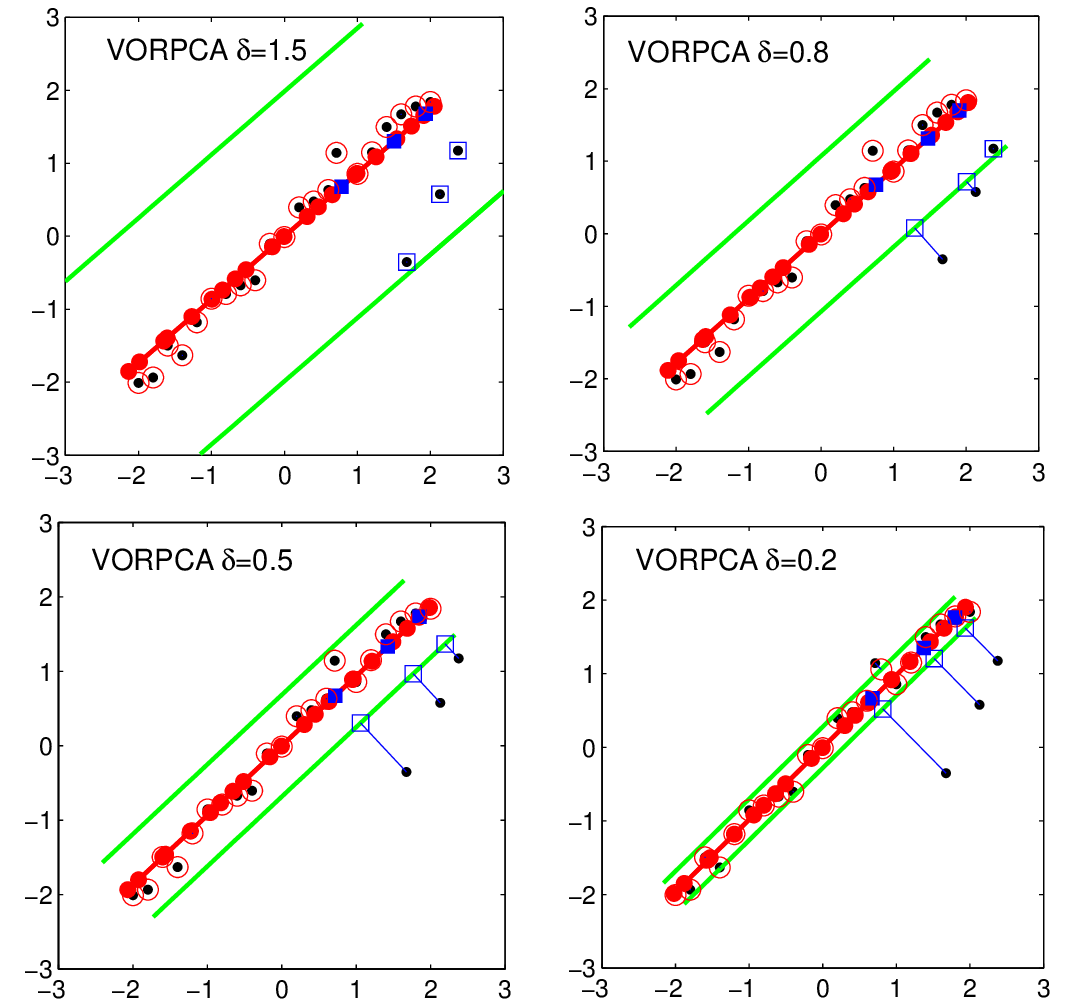}
\caption{
Illustration of corruption tolerant reconstruction using  VORPCA of Eq.(\ref{EQ:ORPCA0}) at four  $\delta$ values.
Original data $\{\textbf{x}_i\}$  are shown in black dots. Reconstructed data $\{ \tilde{\textbf{x}}_i \}$  are shown as red-circles(non-outliers) and red-squares(outliers). Red line indicates prediction correct subspace (manifold). Green lines show the boundary (tolerance limit). Outliers are brought back towards the correct subspace by VORPCA and kept at the boundary (tolerant limit), while non-outliers remain mostly unchanged.}
\end{figure}
\subsection{Computational algorithm}

The above VORPCA problem can be efficiently computed by using the following algorithm.
\newline
\textbf{S0}: Initialize  $\tilde{\textbf{X}} = \textbf{U}\textbf{V}$ .
\newline
Repeat Step \textbf{S1} and \textbf{S2} until convergence.
\newline
\textbf{S1}: Fixing $\textbf{F}=\textbf{U}\textbf{V}$, i.e., $\textbf{f}_i = (\textbf{U}\textbf{V})_{i}$, we compute $\tilde{\textbf{X}}$ using Eq.(\ref{EQ:robust-function})\\
\textbf{S2}: Fixing $\tilde{\textbf{X}}$ in Eq.(\ref{EQ:ORPCA0}), the optimization for $\textbf{U}$ and $\textbf{V}$  is
%
$
\min_{\textbf{U},\textbf{V}} \|\tilde{\textbf{X}}-\textbf{U}\textbf{V}\|^2_F  .
$
%
We minimize $\textbf{U}, \textbf{V}$ alternatively.
Fixing $\textbf{V}$, we compute
\begin{equation}
\textbf{U} = \tilde{\textbf{X}} \textbf{V}^T (\textbf{V} \textbf{V}^T)^{-1}
\label{EQ:L1Reg-PCA5}
\end{equation}
Note $\textbf{V}\textbf{V}^T$ is a $k$-by-$k$ matrix and its inverse is easily computed because $k \simeq 50$.
Fixing $\textbf{U}$, we compute
\begin{equation}
\textbf{V} = (\textbf{U}^T \textbf{U})^{-1} \textbf{U}^T \tilde{\textbf{X}}
\label{EQ:L1Reg-PCA6}
\end{equation}
Here, again,  $\textbf{U}^T\textbf{U}$ is a $k$-by-$k$ matrix and its inverse is easily computed. 
In summary, Eqs.(\ref{EQ:robust-function},\ref{EQ:L1Reg-PCA5},\ref{EQ:L1Reg-PCA6}) form an efficient algorithm to solve Eq.(\ref{EQ:ORPCA0})
This {\it extremely simple} algorithm
is much faster than SVD based algorithms for computing low-rank data approximations.
The convergence of the algorithm is guaranteed because each update has a closed-form solution which decreases the objective function in each iteration.

\subsection{Continuous representation}

Using the continuous representation Eq.(\ref{EQ:robust-function-Con})
of the VOR function,
the VORPCA model
of Eqs.(8, 9)
can be equivalently formulated as
\begin{equation}\label{EQ:ORPCA}
\min_{\textbf{Z},\textbf{U},\textbf{V}} \;  \|\textbf{X}-\textbf{Z}\|_{2,1}+\frac{1}{2\delta} \|\textbf{Z}-\textbf{U}\textbf{V}\|^2_F,
\end{equation}
where the optimal $\tilde{\textbf{X}}$ of Eqs.(8, 9) is the optimal solution $\textbf{Z}$ of problem Eq.(\ref{EQ:ORPCA}).
\begin{figure*}[!htb]\label{2DAPP-L21Reg-PCA}
\centering
\includegraphics[scale=0.55]{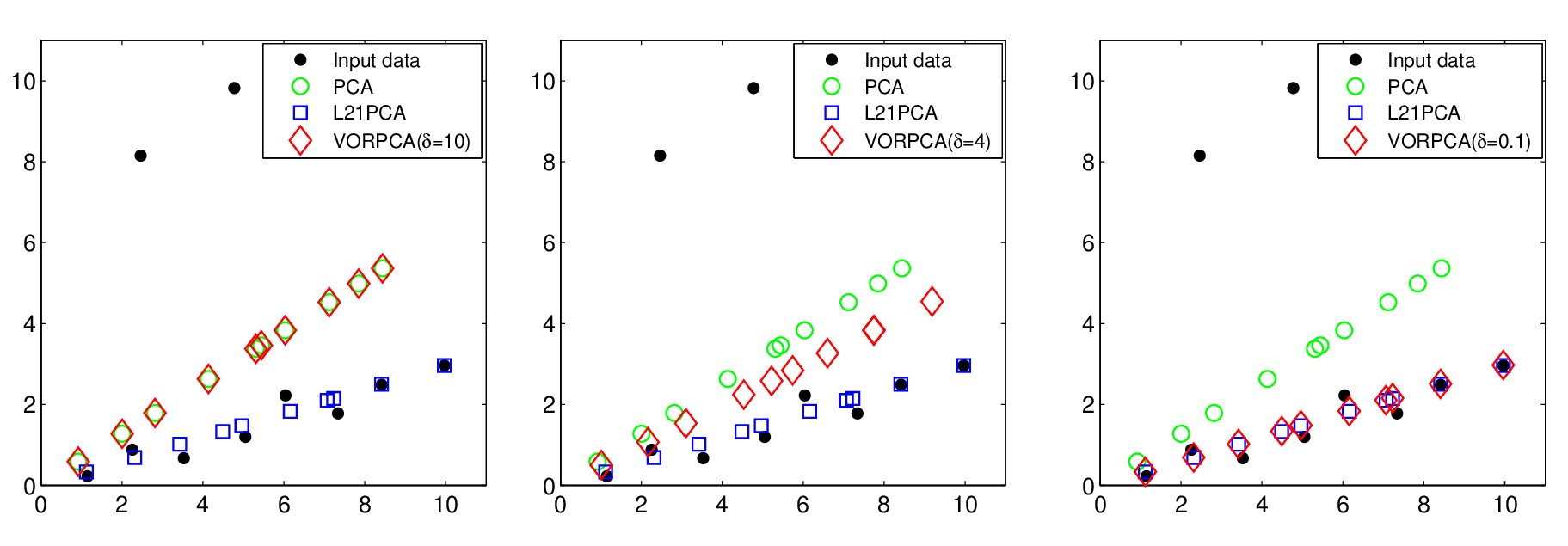}
\caption{
Illustration of VORPCA  at three  $\delta$ values. When $\delta$ is large enough, the reconstruction result of VORPCA is almost identical to standard PCA (L2 norm). When $\delta$ is small enough, the reconstruction result of VORPCA is almost identical to $R_1$-PCA (L21PCA). }
\end{figure*}
\subsection{Connection with PCA and $R_1$-PCA}
In VORPCA model Eq.(\ref{EQ:ORPCA}), when the tolerance  $\delta\to0$, the second reconstruction term is weighted with an infinite weight. Thus $\textbf{Z} = \textbf{U}\textbf{V}$ and VORPCA problem becomes the $L_{2,1}$-norm based PCA ($R_1$-PCA) \cite{R1-PCA,RL1-PCA},
\begin{equation}\label{EQ:R1-PCA}
\min_{\textbf{U},\textbf{V}} \   \| \textbf{X} - \textbf{U}\textbf{V} \|_{2,1}
\end{equation}
On the other hand, when $\delta\to \infty$,  the first term is weighted with  an infinite weight. Thus $\textbf{X} = \textbf{Z}$ and VORPCA problem becomes standard PCA \cite{PatternClassification},
\begin{equation}\label{EQ:R1-PCA}
\min_{\textbf{U},\textbf{V}} \   \| \textbf{X} - \textbf{U}\textbf{V} \|^2_{F}
\end{equation}
Formally, we have
\begin{proposition}
When  $\delta\to0$, VORPCA becomes $R_1$-PCA. When  $\delta\to \infty$, VORPCA becomes PCA.
\end{proposition}
\textbf{Remark.}
Our VORPCA can be regarded as a kind of balanced model between PCA and $R_1$-PCA, as demonstrated in Figure 5.
It has been shown that $R_1$-PCA performs robustly to outliers.
The fact that $R_1$-PCA is the small tolerance limit of VORPCA offers some insights into
$L_{2,1}$ norm robustness.
 At small $\delta$, most data points become outliers and are regularized using the VOR function, i.e, pulled towards the theoretical prediction. Obviously, true outliers do not affect the final results, and furthermore, the outlyingness of true outliers do not matter either.
This provides a new kind of explanation explanation that how robustness are performed in $R_1$-PCA model, i.e., the outliers are corrected using the VOR function.

%

%
%




\section{Comparison with Trace-norm Model}

One advantage of the VORPCA (Eq.(\ref{EQ:ORPCA})) is that it provides both low rank representation $\textbf{Z}$ and also the subspace $\textbf{U}$ and low dimension representation $\textbf{V}$ while eliminates the noises simultaneously. Another important advantage of VORPCA is that the reconstructed data $\textbf{Z}$ does not shrink the magnitude of the data. To show this, we first introduce the trace-norm based PCA which has been widely used in many computer vision and pattern recognition tasks.
The previous work which is closest to
VORPCA is the following trace norm based $L_{2,1}$-PCA (we call it TrL21PCA)~\cite{xu2012robust}:

%
\begin{equation}\label{EQ:TrL21PCA}
\min_{\textbf{Z}} \|\textbf{X}-\textbf{Z}\|_{2,1}+\beta \|\textbf{Z}\|_{tr}.
\end{equation}
Here,
$\beta$ is a positive weighting parameter.
The trace norm (also called nuclear norm)  $\|\textbf{ Z}\|_{tr}$ is the sum of singular values of $\textbf{Z}$). The trace norm
is a surrogate of rank($\textbf{Z}$), with the purpose to achieve low-rank \cite{Fazel:TraceNorm,Recht:TraceNorm}.
One advantage of trace norm is that it is convex. 
In the following, we provide a detail comparison and discussion between our VORPCA of Eq.(\ref{EQ:ORPCA}) and
TrL21PCA model of Eq.(\ref{EQ:TrL21PCA}).
\begin{figure}[!th]\label{AT&T-faceRecL1}
\centering
\includegraphics[scale=0.475]{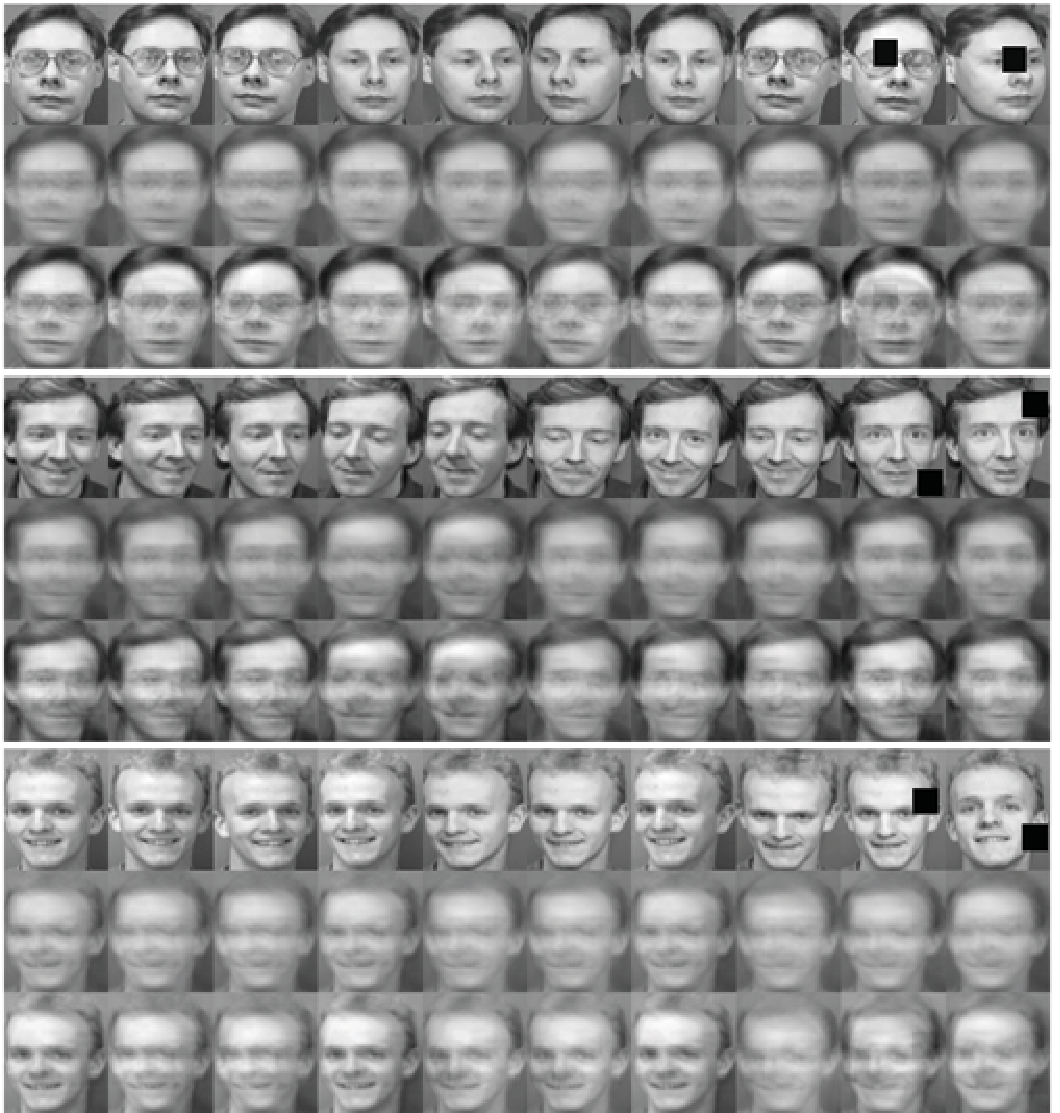}
\caption{Reconstruction results from TrL21PCA and VORPCA on AT\&T face data.
In each panel for images of one person, top line: original occluded images; middle line: reconstruction from TrL21PCA; bottom line: reconstruction from VORPCA.
Finer details of individual images are suppressed in TrL21PCA, but partially retained in VORPCA.}
\end{figure}
\begin{figure*}[!ht]\label{SVD-Curves}
\centering
\includegraphics[scale=0.45]{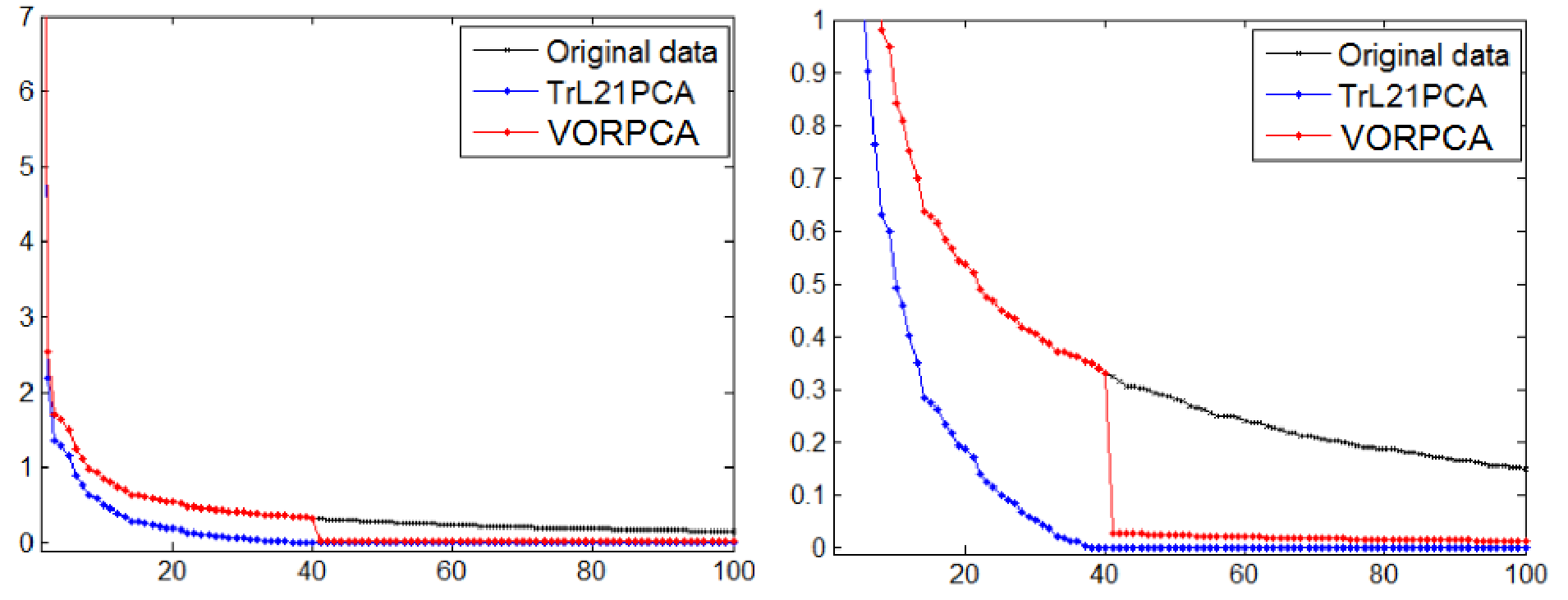}
\caption{
Singular values of solution $Z$ from TrL21PCA and VORPCA on AT\&T face data.
(a) The entire scale.
(b) In small vertical scale such that small singular values are more clear.
The presence of small  high-rank components (with non-zero singular values) in VORPCA help retain
fine details in VORPCA reconstruction.
}
\end{figure*}

%

\subsection{Image reconstruction and noise-free residual}

To help illustrate the main points,
%
we run both VORPCA and TrL21PCA on the occluded images from AT\&T face dataset
(400 images of 10 persons.  More details are given in the Experiments section).
Figure 6 shows image reconstruction comparison for TrL21PCA and VORPCA. Due to space limit, we show only images for 4 persons.
Here we observe that (1) Both VORPCA and TrL21PCA reconstruction are robust w.r.t. large occlusion errors.
(2) Finer details of individual images are
mostly suppressed in TrL21PCA, but are partially retained in VORPCA.
Figure 7 shows singular values of computed $\textbf{Z}$.
Here, one can see that the singular values
 of  TrL21PCA reconstructed data are downshifted (evenly suppressed) for all
 terms. From $k=38$ and up, all singular values are zero.
 In contrast,
 singular values of VORPCA reconstructed data remain close to the original data for ranks $k=1$ to 40,
 but reduce significantly beyond these ranks. They remain non-zero for all higher ranks.
  The sharp singular value drop in VORPCA results near the desired rank
 is the key feature of the proposed model. Small but non-zero higher ranks help retain certain fine details in reconstructed images.


A concise measure of the effects of noise removal can be defined as the following.
Let $\textbf{X}_0$ be original non-occluded images representing true signals. Let $\textbf{E}$ be the
occlusion, i.e., the added noise.
$\textbf{X}=\textbf{X}_0 + \textbf{E}$ is the input data. Let $\textbf{Z}$ be computed from the TrL21PCA and VORPCA models.
Then we define the {\it Noise-free Residual} as
$ \|\textbf{ Z} -\textbf{ X}_0\|_F$.
Figure 8 shows the residual for the faces of different persons. We can see that the VORPCA can usually return lower residual value than TrL21PCA. This is consistent with the Figure 3.
\begin{figure}[!htp]\label{Residual}
\centering
\includegraphics[scale=0.4]{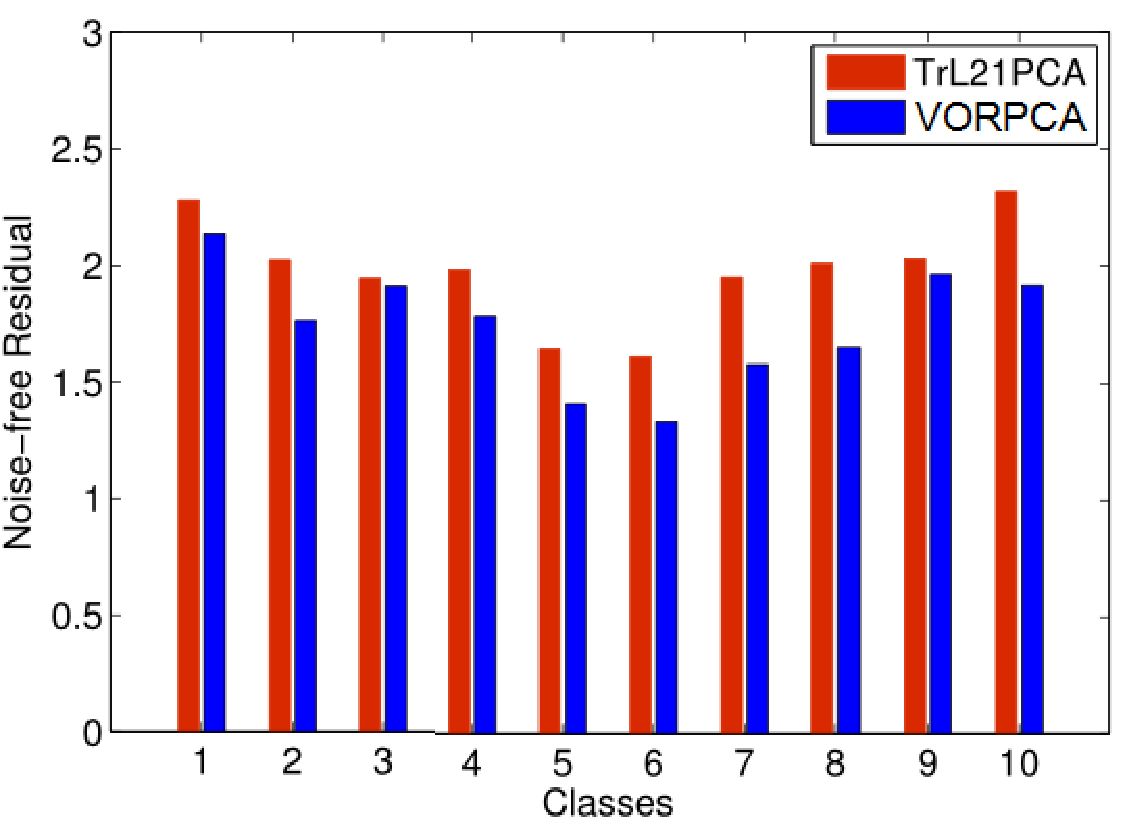}
\caption{Noise-free residual from TrL21PCA and VORPCA on AT\&T data. Each class refers to 10 images of a person. }
\end{figure}
\subsection{Rank suppression and higher rank component}

The trace norm approach Eq.(\ref{EQ:TrL21PCA}) is rank suppression/downshift. Let $\textbf{Z}= \textbf{U}\Sigma \textbf{V}^T$, then  $\| \textbf{Z} \|_{tr} = \sum_k \sigma_k(\textbf{Z})$, where $\sigma_k$ are the singular values of $\textbf{Z}$.
Because this appear directly in the cost function Eq.(\ref{EQ:TrL21PCA}),
thus all rank terms $\sigma_k \textbf{u}_k \textbf{v}_k^T$ are suppressed. 
Especially, due to the uniform downshift of singular values, higher rank terms are completed suppressed, as can be seen from Figure 7.
The uniform singular value downshift
can also be seen from the solvable case \cite{Ma:TraceNorm}
\begin{equation}\label{EQ:L2TPCA}
\min_{\textbf{Z}} \|\textbf{ X}-\textbf{Z} \|_F^2 + \beta \|\textbf{Z}\|_{tr}.
\end{equation}
which has closed form  solution:
$\textbf{Z}^* = \textbf{U}(\Sigma-\beta \textbf{I})_+\textbf{V}^T$. Note that the diagonal factor $(\Sigma-\beta \textbf{I})_+$ uniformly downshift all singular values.

In contrast, our VORPCA model has the following two aspects: (1) In VORPCA, the data rank are only suppressed on higher rank terms.
The important lower ranks $1 \le k \le K$ are not suppressed, but instead {\it protected}.
This can be seen from Figure 4 ($K$ = 40) where lower rank singular values remains nearly identical to input data.
(2) In VORPCA, higher rank components do not appear directly in cost function. They suppressed, but not completely eliminated,
as can be seen from Figure 7(b). Small but non-zero higher ranks help retain certain fine details in reconstructed images.
This can be seen in Figure 6.

\section{Experiments}

We run the proposed VORPCA model and
compare with other models 
on several image datasets~\footnote{They are available from \\ http://www.cad.zju.edu.cn/home/dengcai/Data/data.html}, including AT\&T, Bin-alpha, MNIST, USPS and COIL20~\cite{nene1996columbia}.
We perform clustering task on different datasets and compare clustering results on eight data representations:
(1) original data,
(2) standard PCA~\cite{PatternClassification},
(3) TrL21PCA~\cite{xu2012robust},
(4) Robust PCA (RPCA)~\cite{RPCA},
(5) Laplacian Embedding (LE)~\cite{LE},
(6) Normalized cut (Ncut)~\cite{shi2000normalized},
(7) VORPCA (Z),
(8) VORPCA (V).
 In our VORPCA, we can either work directly on $\textbf{Z}$ which has the same dimension as the original data.
 This is the version Z above. We can also work on $\textbf{V}$ which has much smaller dimension of $k$. This is version V above.
 We use K-means clustering for this evaluation. We run K-means with random initialization 50 times and use the average clustering result. Results are shown in Table I.  
From Table I, we  observe that (1) PCA performs poorly on occluded data (AT\&T), indicating that PCA is sensitive to outlying observation. (2) Both TrL21PCA and VORPCA perform in a similar manner and return better performance than other alternatives on all datasets. This suggests that both TrL21PCA and VORPCA are robust to the outliers. (3) VORPCA(V) generally performs better than other methods.

\begin{table*}[htpb!]
	\center
	\caption{Clustering results on five datasets}\label{tabel2}
	\begin{tabular}{c|c|c|c|c|cr}
		\hline
		\hline
		&AT\&Tocc &USPS &MINIST  &BinAlpha &COIL  \\
		\hline
		Original	&0.6330 	&0.5795	  &0.5117	  &0.5212	&0.5727 \\
		\hline
		PCA	    &0.6395 	&0.5870	  &0.5252	  &0.5038	&0.5881	  \\
		\hline
		TrL21PCA	&0.6610 	&0.6060	  &0.5408	  &0.5393	&0.6144	\\
		\hline
		RPCA	&0.6581     &0.6199	  &0.5390	  &0.5593	&0.5909	    \\
		\hline
		LE	    &0.6191	    &0.5662	  &0.5489	  &0.5167	&0.5979	    \\
		\hline
		Ncut	  &0.6637	    &0.5809	  &0.5511	  &0.5667 &0.5675	\\
		\hline
		VORPCA(Z)	&0.6891	   &0.6333	  &0.5625	  &0.5787 &0.6290	\\
		\hline
		VORPCA(V)	&0.6898	    &0.6553	  &0.5640	  &0.5645 &0.6257	\\
		\hline	
		\hline
	\end{tabular}
\end{table*}
%


We perform clustering task on different datasets and compare clustering results on eight data representations:
(1) original data,
(2) standard PCA,
(3) TrL21PCA,
(4) Robust PCA (RPCA),
(5) Laplacian Embedding (LE),
(6) Normalized cut (Ncut),
(7) VORPCA (Z),
(8) VORPCA (V).
 In our VORPCA, we can either work directly on $\textbf{Z}$ which has the same dimension as the original data.
 This is the version Z above. We can also work on $\textbf{V}$ which has much smaller dimension of $k$. This is version V above.
 We use K-means clustering for this evaluation. We run K-means with random initialization 50 times and use the average clustering result. Results are shown in Table 2. Clustering accuracy are computed as the known class labels. This is done as follows: the confusion matrix is first computed. The columns and rows are then reordered so as to maximize the sum of the diagonal. We take this sum as a measure of the accuracy: it represents the percentage of data points correctly clustered under the optimized permutation. 
From Table 2, we  observe that (1) PCA performs poorly on occluded data (AT\&T), indicating PCA is sensitive to outlying observation. (2) Clustering in the both TrL21PCA and VORPCA(Z) performs in a similar manner and returns better performance than other alternatives on all the datasets. This suggests that both TrL21PCA and VORPCA are robust to the gross noise. (3) VORPCA(V) generally performs better than other data representations.

\section{Conclusions}

In this paper, we
introduce {\it vector outlier regularization} (VOR) function. VOR provides a kind of intuitive explanation for $L_{2,1}$ norm robustness w.r.t outliers.
We use the VOR function to construct VORPCA and
present an efficient algorithm to compute VORPCA and demonstrate its robustness.
We provide theoretical analysis and continuous formation of VORPCA.
We provide theoretical analysis and continuous formulation of VORPCA to demonstrate the robustness of $R_1$-PCA model.

%
%

\section*{Acknowledgment}
This work is supported by
 National Natural Science
Foundation of China (61602001, 61572030); Open fund for Discipline Construction, Institute of Physical Science and Information Technology, Anhui University

\section*{Appendix}

We prove that the solution of Eq.(\ref{EQ:proximal}) is given by
Eq.(\ref{EQ:robust-function4}).
We recently note that similar proofs have been provided in works~\cite{L21F,L20}.
Here, we provide another simple proof.

To simplify the notation, we ignore the subscript $i$ in
Eq.(\ref{EQ:proximal}),
and write it as


\begin{equation}\label{EQ:robust-function8}
\min_{\textbf{u}} \delta \|\textbf{u}\|+ \frac{1}{2} \|\textbf{u} -\textbf{a} \|^2.
\end{equation}
\textbf{Proof.} It is clear that, given the magnitude of the vector $\textbf{u}$,
the direction of $\textbf{u}$
must be in the same direction of the vector $\textbf{a}$ in order to minimize the
second term.
Thus
the direction of $\textbf{u}$
must be in the same direction of the vector $\textbf{a}$, i.e,
we must have $ \textbf{u} = \rho \textbf{a}$ where $\rho \ge 0$ is a scalar.

Substituting to Eq.(\ref{EQ:robust-function8}), we need to minimize
$$
f(\rho) = \delta \| \textbf{a} \| \rho + (1/2) \| \textbf{a} \|^2 (\rho - 1)^2.
$$
subject to $\rho \ge 0$.
The KKT complementarity slackness condition~\cite{nocedal2006numerical} is
$$
0 = \rho f'(\rho) = \rho [\delta + (\rho-1)\| \textbf{a} \| \ ].
$$
The solution is $\rho^* = \max (1 - \delta/\| \textbf{a} \|, 0)$. This gives $\textbf{u}^*
= \rho^* \textbf{a} $.
Replacing $\textbf{a} = \textbf{f}_i - \textbf{x}_i$,
this gives Eq.(\ref{EQ:robust-function4}),
This completes the proof.

\bibliographystyle{named}
\bibliography{example_paper2}

\end{document}